\documentclass[runningheads]{llncs}
\usepackage{graphicx}
\usepackage{multirow}
\usepackage{todonotes}
\usepackage{stmaryrd}
\usepackage{ifthen}
\usepackage{xspace}
\usepackage{amsmath}
\usepackage{amssymb}

\newcommand{\textmacro}[2]{\newcommand{#1}{#2\xspace}}

\newcommand{\automl}{AutoML}
\newcommand{\autoweka}{Auto-WEKA}
\newcommand{\autoscikit}{auto-sklearn}
\renewcommand{\vec}[1]{\boldsymbol{#1}}

\newboolean{long}
\setboolean{long}{false}
\textmacro{\tool}{ML$^2$-Plan}

\begin{document}
\title{Automated Multi-Label Classification\\ based on ML-Plan}
%
%
\author{Marcel Wever \and
Felix Mohr \and
Eyke H{\"u}llermeier}
\authorrunning{M. Wever et al.}
%
\institute{Heinz Nixdorf Institute\\
Paderborn University, Germany\\
\email{marcel.wever@upb.de,felix.mohr@upb.de,eyke@upb.de}\\
}

\maketitle              
\begin{abstract}
Automated machine learning (\automl{}) has received increasing attention in the recent past.
While the main tools for \automl{}, such as \autoweka, TPOT, and \autoscikit{}, mainly deal with single-label classification and regression, there is very little work on other types of machine learning tasks.
In particular, there is almost no work on automating the engineering of machine learning applications for multi-label classification.
This paper makes two contributions.
First, it discusses the usefulness and feasibility of an \automl{} approach for multi-label classification.
Second, we show how the scope of ML-Plan, an \automl-tool for multi-class classification, can be extended towards multi-label classification using MEKA, which is a multi-label extension of the well-known Java library WEKA.
The resulting approach recursively refines MEKA's multi-label classifiers, which sometimes nest another multi-label classifier, up to the selection of a single-label base learner provided by WEKA.
In our evaluation, we find that the proposed approach yields superb results and performs significantly better than a set of baselines.

\end{abstract}
\section{Introduction}
\label{sec:introduction}
These days, machine learning functionality is required in more and more application areas, and machine learning applications have already become part of everyday life.
Since end users in application domains are normally not machine learning experts, there is an urgent need for suitable support in terms of tools that are easy to use.
Ideally, the induction of models from data, including the data preprocessing, the choice of a model class, the training and evaluation of a predictor, the representation and interpretation of results, etc., would be automated to a large extent \cite{LloydDGTG14}.
This has triggered the field of \emph{automated machine learning} (\automl), which has developed into an important branch of machine learning research in the last couple of years. 

State-of-the-art \automl{} tools \cite{autoweka,komer2014hyperopt,feurer2015efficient} have shown impressive results on multi-class classification problems.  
These approaches are essentially based on a formalization of the \automl{} problem in terms of an optimization problem with a fixed number of decision variables, amenable to standard (Bayesian) optimization tools such as SMAC.
Typically, there is one variable for the preprocessing algorithm, one variable for the learning algorithm, and one variable for each parameter of each algorithm.
While this technique works well for problems with no or only little hierarchical structure, it is less suitable for more complex problems whose solutions are naturally designed in a recursive manner.

An example of such a problem is multi-label classification (MLC), which is the topic of this paper. One reason for the natural appearance of recursion in MLC is the common use of meta-learning techniques for reducing multi-label to binary or multi-class problems. 
Almost each such learner takes a base learner as input, which, in principle, could be an entire machine learning pipeline (ML pipeline) itself.
However, there has been very little work on \automl{} for multi-label classification (MLC), i.e., finding good multi-label classifiers in an automated fashion.
Besides the work of de S\'a et al. \cite{DBLP:conf/gecco/SaPF17,DBLP:conf/ppsn/SaFP18} based on genetic algorithms, we are not aware of previous work on automated multi-label classification.

In this paper, we discuss the usefulness and the feasibility of an \automl{} approach for MLC.
As multi-label classifiers usually reduce the MLC task to several single-label classification tasks, the configuration of a multi-label classifier is of a hierarchical nature. Therefore, we propose to use a \emph{hierarchical} technique for the configuration of such ML pipelines, which caters more naturally for the hierarchical structure of the problem.
Starting with an empty pipeline, the algorithm first selects a multi-label classifier. If the chosen multi-label classifier represents a meta strategy, it is refined with another multi-label classifier. Otherwise, multi-label classifiers usually require a base learner for binary classification, which in turn has to be selected from a portfolio of single-label classifiers.

To configure these multi-label classifiers, we adapt and use an \automl{} tool for single-label classification called ML-Plan \cite{mlplan}.
ML-Plan leverages a derivative of hierarchical task network (HTN) planning \cite{GeorgievskiA15}, called programmatic task network planning \cite{ptnplanning}, to solve the \automl{} task, and thus, it naturally supports recursive structures as they appear in \automl{} for multi-label classification.
For instance, in \cite{mlplanUnlimited} it is shown how ML-Plan can produce tree-shaped preprocessing workflows of arbitrary depth.
This is also the reason for why we prefer ML-Plan to other \automl{} tools. In the following, we refer to this adapted version of ML-Plan as \tool{} (Multi-Label ML-Plan).
We empirically show that this approach performs particularly well and significantly outperforms the baselines.
Due to the lack of dedicated \automl{} tools for the task of multi-label classification (except the recent work of \cite{DBLP:conf/gecco/SaPF17} based on genetic algorithms), a comparison is not straight-forward. Nevertheless,
we managed to set up meaningful and reasonably strong baselines including a \ifthenelse{\boolean{long}}{grid-search, }{}random search, and a reduction to a single-label classification \automl{} tool.

\section{Multi-Label Classification}
\label{sec:problem}
In contrast to conventional (single-label) classification, the setting of \emph{multi-label classification} (MLC) allows an instance to belong to several classes simultaneously, i.e., to be assigned several labels at the same time. For example, a single image could be tagged simultaneously with labels \texttt{Sun} and \texttt{Beach} and \texttt{Sea}. 

More formally, let $\mathcal{X}$ denote an instance space, and let $\mathcal{L}= \{\lambda_1, \lambda_2, \ldots, \lambda_m\}$ be a finite set of class labels.
We assume that an instance $\vec{x} \in \mathcal{X}$ is (non-deterministically) associated with a subset of labels $L \in 2^\mathcal{L}$; this subset is often called the set of relevant labels, while the complement
$\mathcal{L} \setminus L$ is considered as irrelevant for $\vec{x}$.
We identify a set $L$ of relevant labels with a binary vector $\vec{y} = (y_1, y_2, \ldots, y_m)$, in which $y_i = 1$ iff $\lambda_i \in L$. By $\mathcal{Y} = \{0,1\}^m$ we denote the set of possible labelings. 

In general, a multi-label classifier $\mathbf{h}$ is a mapping $\mathcal{X} \rightarrow \{0,1\}^m$. For a given instance $\vec{x}\in \mathcal{X}$, it returns a prediction in the form of a vector
\[
\mathbf{h}(\vec{x}) = \big(h_1(\vec{x}), h_2(\vec{x}), \ldots , h_m(\vec{x}) \big)\,.
\]
The problem of MLC can be stated as follows: Given training data in the form of a finite set of observations
$$
\big\{ (\vec{x}_i,\vec{y}_i) \big\}_{i=1}^N \subset \mathcal{X} \times \mathcal{Y \, }, 
$$
the goal is to learn a classifier $\mathbf{h}: \, \mathcal{X} \rightarrow \{0,1\}^m$ that generalizes well beyond these observations in the sense of minimizing the risk with respect to a specific loss function.

There are various loss functions that are commonly used in MLC, including the subset 0/1 loss (exact match)\footnote{$\llbracket \cdot \rrbracket$ is the indicator function.}
\[
L_{0/1}(\vec{y}, \mathbf{h}(\vec{x})) = 
\llbracket \vec{y} \neq \mathbf{h}(\vec{x}) \rrbracket \enspace ,
\]
the Hamming loss
\[
L_{H} (\vec{y}, \mathbf{h}(\vec{x})) =
\frac{1}{m} \sum_{i=1}^m \llbracket y_i \neq h_i(\vec{x})) \rrbracket \enspace ,
\]
and the (instance-wise) F-measure (which is actually a measure of accuracy)
\begin{equation}\label{eq:fmeasure}
F (\vec{y}, \mathbf{h}(\vec{x})) =
\frac{2 \sum_{i=1}^m y_i h_i(\vec{x})}{\sum_{i=1}^m y_i + \sum_{i=1}^m h_i(\vec{x}) }
\enspace .
\end{equation}
In slightly different tasks like ranking or probability estimation, the prediction of a classifier is not restricted to binary vectors. Instead, a hypothesis $\mathbf{h}$ is a mapping $\mathcal{X} \rightarrow \mathbb{R}^m$, which assigns scores to labels. Corresponding predictions also require other loss functions. An example is the rank loss, which compares a ground-truth labeling with a predicted ranking of the labels and counts the number of incorrectly ordered label pairs: 
\[
L_R(\vec{y},\mathbf{h}(\vec{x})) =  \sum_{(i,j): y_i > y_j}  \left (\llbracket h_i  (\vec{x}) < h_j(\vec{x}) \rrbracket +  
 \frac{1}{2} \llbracket h_i(\vec{x}) = h_j(\vec{x}) \rrbracket  \right)
\]

At first sight, MLC problems can be solved in a quite straightforward way, namely through decomposition into several binary classification problems: One binary classifier is trained for each label and used to predict whether, for a given query instance, this label is present (relevant) or not. This approach is known as \emph{binary relevance} (BR) learning.

However, BR has been criticized for ignoring important information hidden in the label space, namely information about the interdependencies between the labels. Since the presence or absence of the different class labels has to be predicted \emph{simultaneously}, it is arguably important to exploit any such dependencies.

Going beyond BR, a large repertoire of methods for MLC has been proposed in the recent years. Most of these methods seek to improve predictive accuracy by exploiting label dependencies in one way or the other. We refer to \cite{DBLP:journals/tkde/ZhangZ14} for an up-to-date survey on MLC algorithms.

\section{\automl\ and Hierarchical Planning}
\label{sec:approach}

\automl{} seeks to automatically \emph{compose} and \emph{parametrize} machine learning algorithms into ML pipelines, with the goal to optimize a given metric, e.g., minimizing the exact match loss or maximizing the instance-wise F-measure in the case of MLC.
The algorithms are typically related either to preprocessing (feature selection, transformation, imputation, etc.) or to the core functionality (classification, regression, ranking, etc.).

While there is no general limitation on the structure of the composition of these algorithms (they are unbound in length and may contain alternative branches or even loops), the pipelines created by current approaches are usually rather simple and essentially limited to a preprocessing step and a classifier.
For ML problems more complex than standard classification, approaches of that kind are not fully suitable.

\begin{figure}[t]
	\includegraphics[width=0.49\columnwidth]{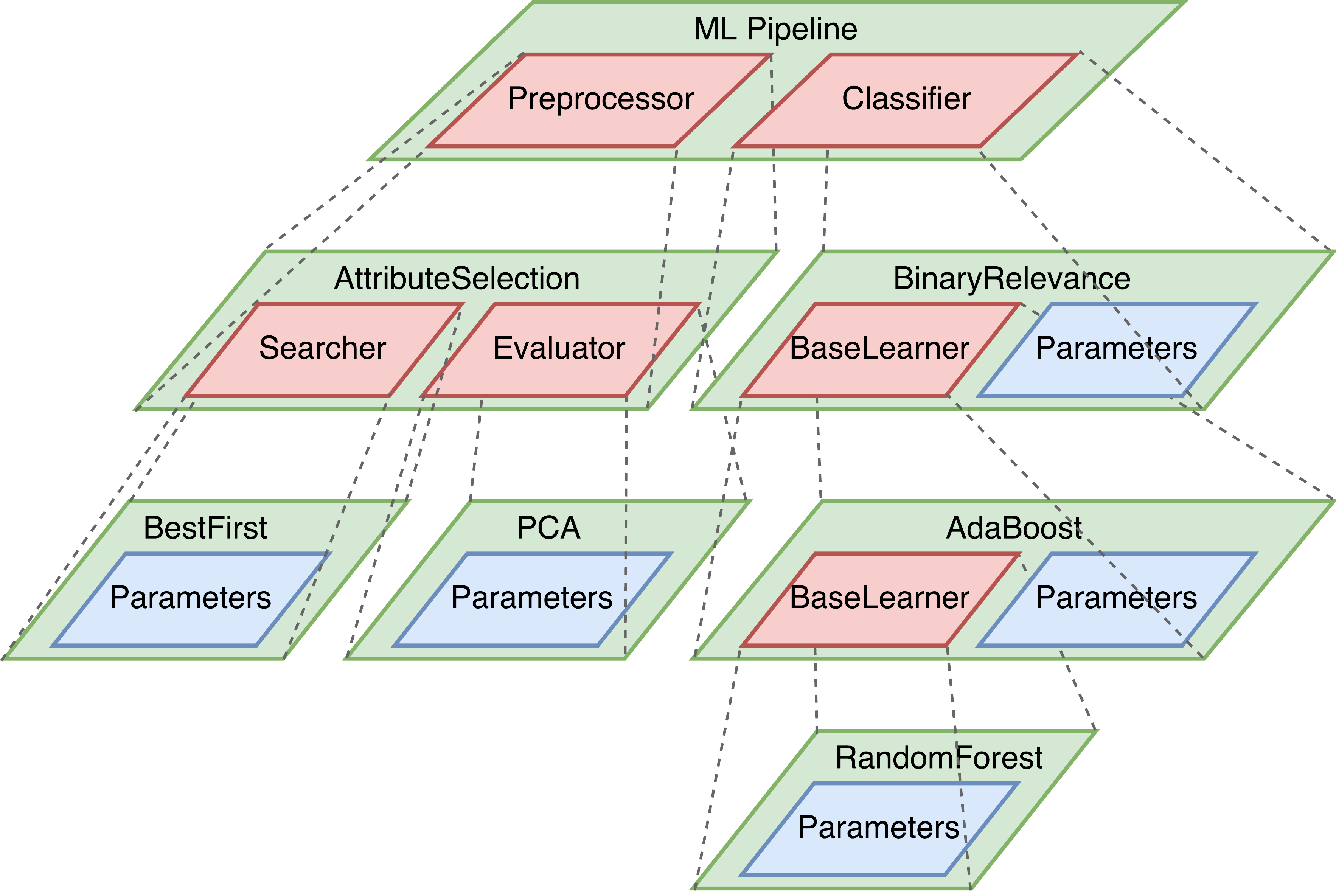}
	\hspace{.1em}
	\includegraphics[width=0.49\columnwidth]{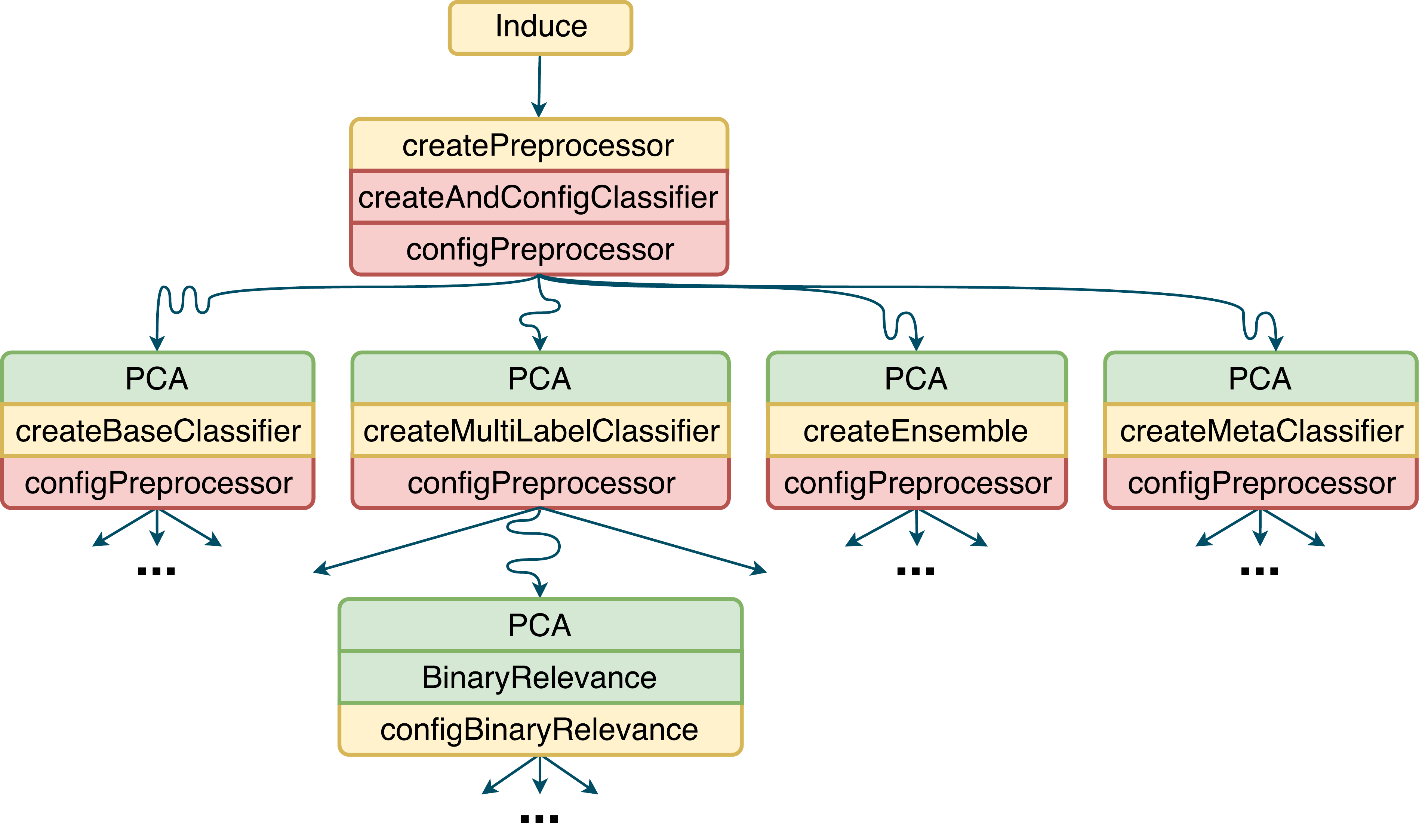}
    \label{fig:decomposition}
    \caption{Visualization of the hierarchical structure of a machine learning pipeline (left) and an excerpt of the hierarchical planning search graph (right).}
\end{figure}

The decomposition scheme on the left-hand side in Fig. \ref{fig:decomposition} suggests that we can construct a machine learning pipeline in a hierarchical way.
In an initial step, we have the complex tasks to choose a preprocessor (possibly an empty one) and a classifier.
However, each of these components may need other components and/or parameters in turn.
So we need to choose and configure these sub-components and parameters first, which, as illustrated in the figure, may have sub-components and parameters, too.
This recursion continues until no more refinement is necessary or possible.

In this paper, to create multi-label classifiers, we make use of hierarchical planning as a formalism that is amenable to this recursive structure.
Hierarchical planning is a concept from the field of AI planning \cite{ghallab2004automated}.
The core idea is to iteratively break down an initially given complex task into new sub-tasks, which may also be complex or simple (no need of further refinement).
The complex tasks are recursively decomposed until only simple tasks remain.
This is comparable, for example, to deriving a sentence from a context-free grammar, where complex tasks correspond to non-terminals and simple tasks are terminal symbols.

Note that there is not one canonical but many possible hierarchical planning problems that can be used to hierarchically construct pipelines.
Just to give an example, we could first choose the algorithms for the multi-label classifier and set their parameters, and then choose the single-label classifier as a base learner, its sub-components, and their parameters.
Yet, we may also nest the process of choosing algorithms and setting the parameters, e.g., to first choose the multi-label classifier and the binary classifier algorithms, and \emph{then} set the parameters of both of them.
While this looks like a trivial change that does not affect the set of pipelines that can be constructed, it has dramatic effects on the structure of the search tree.

Algorithmically, a planning problem is solved using graph search algorithms.
The (hierarchical) planning problem induces a (possibly infinite) search graph, which is represented by a distinguished root node, a successor generator function, and a goal-test function.
The successor generator creates the successor nodes for any node of the graph, and the goal-test decides whether a node is a goal.
Most HTN planners perform a \emph{forward-decomposition}, which means that they create one successor for each possible decomposition of the first unsolved task in the list of remaining tasks.
In every child node, the list of remaining tasks is the previous list of tasks where the decomposed task is replaced by the list that represents the respective decomposition.
The resulting search graph is sketched on the right-hand side in Fig. \ref{fig:decomposition} where every box shows a list of tasks (green ones are simple, the yellow one is the next complex task to be decomposed, and the red ones are complex to be resolved later).
A node is a goal node if all remaining tasks are simple.
A standard graph search algorithm can then be used to identify a path from the root to a goal node, which induces a solution plan.

However, it is not easily possible to solve \automl{} problems using standard planners such as SHOP2 \cite{DBLP:journals/corr/abs-1106-4869}.
The main problem is that, in contrast to the usual assumption of standard planners applying A* search, the cost of a solution (e.g., expected loss of a classifier) cannot be computed from the descriptions of the plan elements. 

We are aware of three approaches to \automl{} using hierarchical planning or related techniques.
The first approach is related to optimization within the RapidMiner framework based on hierarchical task networks (HTN) \cite{DBLP:journals/jair/NguyenHK14,DBLP:conf/ecai/KietzSBF12}.
They conduct a beam search (hill-climbing in the most extreme case), where the beam is selected based on a \emph{ranking} of alternative choices obtained from a meta-learning module, which compares the current dataset with previous ones and choices taken back then.
The most recent representative of this line of research is Meta-Miner \cite{DBLP:journals/jair/NguyenHK14}.
While these approaches do not execute candidates during search to observe their performance, an approach of extensive evaluation is presented in RECIPE \cite{DBLP:conf/eurogp/SaPOP17}.
RECIPE creates pipelines using a grammar-based genetic algorithm; the pipeline candidates are evaluated in the course of computing their fitness.
Third, ML-Plan \cite{mlplan} recognizes the value of executing pipelines during search, but also observes that the extensive evaluation conducted in TPOT and RECIPE is infeasible for larger datasets.
It reduces the number of evaluations by only considering candidates obtained from completions of currently best candidates.
Like Meta-Learner, it is based on HTN planning.

While none of the above approaches has been used to solve multi-label classification problems, they can be adapted into that direction rather easily.
This is precisely thanks to the hierarchical view on the solution candidates, because pipelines for MLC are very similar to pipelines for multi-class classification.
The main difference between the pipelines is that there is at least one more layer of recursion in the configuration of its elements, which is easily incorporated within a hierarchical model.
The adaptation for the HTN-based approaches is particularly appealing, because a huge part of the search graph definition (the one concerned with the configuration of base classifiers) can be simply adopted without any changes.

Of course, other \automl{} solutions such as \autoweka{} or \autoscikit{} can be extended to the multi-label problem as well.
It is clear that one can flatten any clearly limited hierarchical structure into a vector as long as the allowed structures are bound in length, which makes those approaches generally applicable.
In fact, this was already done in both frameworks to cope with preprocessors and meta classifiers.

In this paper, we solve the multi-label \automl{} problem by extending ML-Plan, our approach to automated multi-class classification.
The resulting algorithm is called \tool, which stands for Planning for Multi Label Machine Learning.
In the context of ML-Plan, the extension is quite straight-forward and essentially comes down  to augmenting the already existing hierarchical planning problem definition by the MLC algorithms.
In the following section, we give a brief overview of ML-Plan and how it is extended to \tool.

\section{A Multi-Label Version of ML-Plan}
\subsection{ML-Plan}
As briefly sketched above, ML-Plan is a hierarchical planner designed for \automl{} problems \cite{mlplan}.
Standard hierarchical planners such as SHOP2 \cite{DBLP:journals/corr/abs-1106-4869} lack some fundamental requirements of \automl, e.g., to evaluate candidate solutions during search, which was a main motivation for developing ML-Plan.

The search technique adopted by ML-Plan is a best-first search.
ML-Plan makes no assumption (like monotonicity) about the node evaluations or how they are acquired.
Instead, ML-Plan simply requires that the node evaluation function is provided by the user.
It is then possible to conduct complex computations in order to obtain node evaluations, a property that is missing in classical planners.
The node evaluation in ML-Plan is based on random path completion as also used in Monte Carlo Tree Search \cite{DBLP:journals/tciaig/BrownePWLCRTPSC12}.
To obtain the evaluation of a node, this strategy draws a fixed number of path completions, builds the corresponding pipelines and evaluates them against a validation set.
The score assigned to the node is the \emph{minimal} score that was observed over these validations in order to estimate the best solution that can be obtained when following paths under the node.

\begin{figure}[t]
    \centering
	\includegraphics[width=0.5\columnwidth]{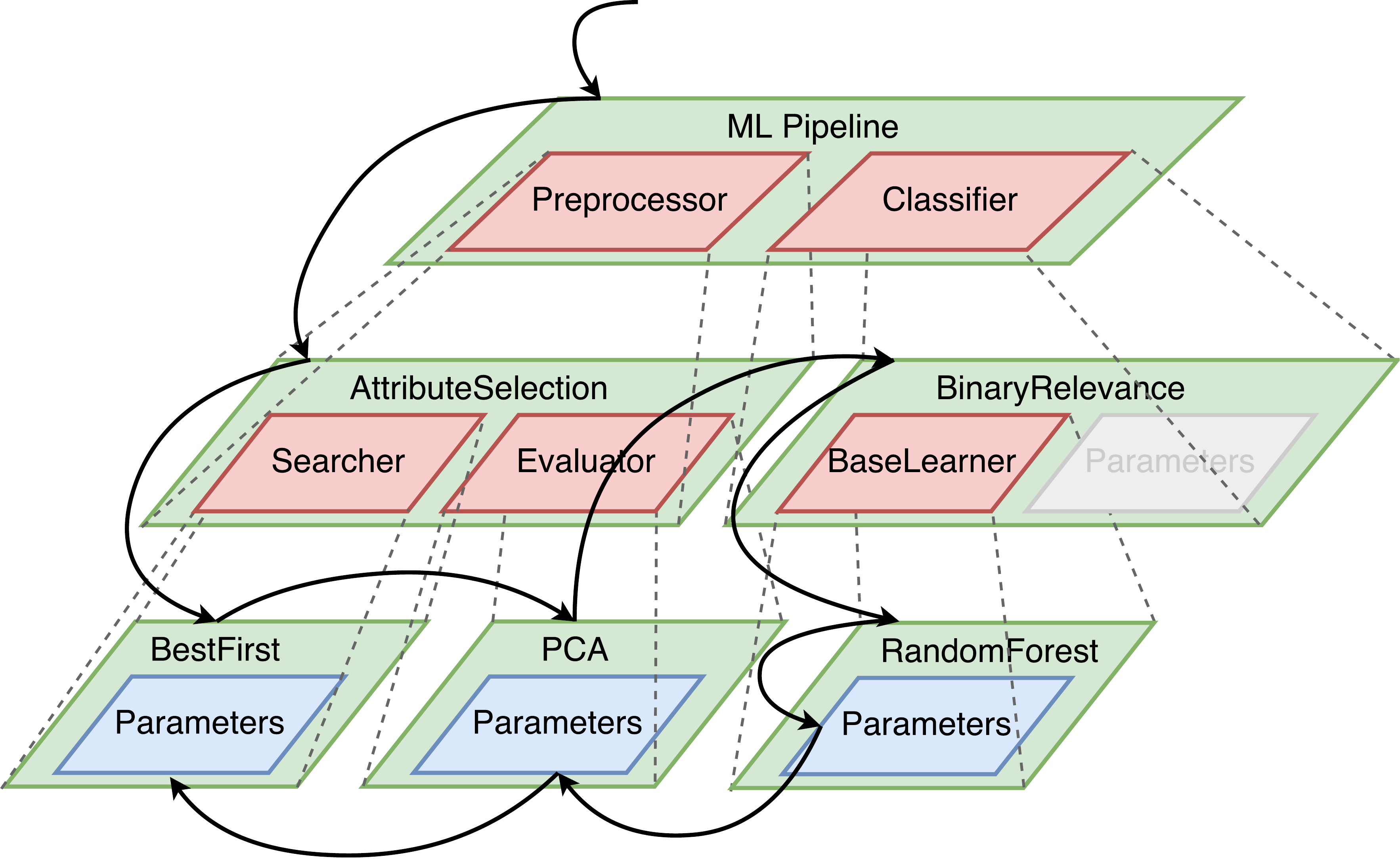}
    \caption{Process of hierarchically refining an ML pipeline}
    \label{fig:decomposition-hops}
\end{figure}

Intuitively, ML-Plan formalizes the HTN problem in a way that the resulting search graph is split into an algorithm selection region (upper region) and an algorithm configuration region (lower region).
This idea is captured in Fig.\ \ref{fig:decomposition-hops}.
The main motivation for this strategy lies in the node evaluation we want to apply, which is based on random completions.
Since algorithm selections usually constitute a much more significant change to the performance of a pipeline than parameter settings, we consider all solutions under a node that has all algorithms fixed as a kind of neighborhood, and the random samples drawn in that lower region are then more reliable estimates.

Having the idea of a two-phased search graph in mind, the HTN definition of ML-Plan is as follows\footnote{Since we have not formally introduced HTN planning, we describe the problem definition in a rather intuitive way.
The formal definition can be found in the implementation published with this paper}.
The initial task \texttt{createClassifier} can be broken down into a chain of the three tasks \texttt{createRawPP}, \texttt{setupClassifier}, \texttt{refinePP}.
The first task is meant to choose the algorithms used for pre-processing \emph{without} parametrizing them, the second task is meant to choose and configure the multi-label classifier, and the third step parametrizes the previously chosen pre-processors.
The second task \texttt{setupClassifier} can, for each classifier, be decomposed into two sub-tasks.
First, \texttt{<classifier>:create} is a simple task indicating the creation of a new classifier of the respective class, e.g.\ \texttt{J48:create}.
Second, \texttt{<classifier>:configure} is a complex task meant to configure the parameters of the classifier.

As an additional remark, ML-Plan comes with a built-in strategy to prevent over-fitting.
This strategy apportions the assigned timeout for the whole search process among two phases.
The first phase covers the actual search in the space.
The second phase takes a collection of identified solutions and selects the one that minimizes the estimated generalization error.
Roughly speaking, the collection used for selection in phase 2 corresponds to the $k$ best candidates and $k$ random candidates that are not significantly worse than the best candidate.
The time allocated at time step $t$ for the second phase is flexible and corresponds to the accumulated time that was required in phase 1 to evaluate the classifiers that would be chosen at time step $t$ for the selection process.

\subsection{Deriving \tool{} from ML-Plan}
To obtain \tool{} from ML-Plan, we need to make two changes.
First, we modify the HTN problem definition to support MLC algorithms and omit tasks that are not necessary or reasonable in MLC.
Second, we adjust the node evaluation function to be based on multi-label loss functions.
We now explain these two aspects in more detail.

\tool modifies the HTN planning problem of ML-Plan in two ways:
\begin{enumerate}
    \item It \emph{simplifies} the problem by removing preprocessing and by deactivating parameter configuration, so \tool only conducts algorithm \emph{selection}.
    Preprocessing is ignored because the preprocessors of multi-class classification are not directly applicable to the multi-label classification case; there are extensions \cite{DBLP:conf/ai/GharroudiEA14,DBLP:journals/air/PereiraPZM18}, but no implementations are available in the used libraries.
    Algorithm configuration is ignored, because the evaluations of solution candidates are usually so expensive that, in the current form, even the algorithm selection problem cannot be solved within a reasonable time bound.
    
    \item It \emph{extends} the problem by adding new algorithms, the MLC algorithms, and by introducing a dedicated notation for the decisions on dependent sub-classifiers.
    The latter is to overcome the previous practice that sub-classifiers are seen as parameters.
    Now, there is a dedicated task for each sub-algorithm of each algorithm.
    For example, the meta-classifiers (both multi-label and multi-class) need to be refined by choosing their base classifiers.
    As single-label classifiers \emph{are} the sub-classifiers of most basic multi-label classifiers, up to three recursions are possible in this way.
    
\end{enumerate}

Technically, the initial task now becomes \texttt{createMLClassifier} to indicate that a multi-label classifier needs to be constructed.
It can be resolved either by a meta multi-label classifier, which results in a new task \texttt{createMLBaseClassifier} or a simple multi-label classifier, which already solves the task in case of the majority classifier and induces a task \texttt{createWekaClassifier} for the configuration of the used base learner in any other case.
This task, which roughly corresponds to \texttt{setupClassifier} in ML-Plan, can be either refined to an empty rest problem by choosing a basic learner like decision trees or a neural network, or it can be refined with a meta learner like AdaBoost, which induces a further task \texttt{setupBaseClassifier}; the latter task can then only be refined with non-meta single-label classifiers.

The set of applied algorithms is a strict superset of the ones adopted in ML-Plan.
The number of possible candidates of composed multi-label classifiers is roughly 80,000. Thus, the space of possible candidates is \emph{much} smaller than in the case of multi-class classification where hyperparameter optimization is performed as well.
However, due to the much costlier evaluations per pipeline, the traversion of a bigger search space does not appear reasonable.
The learners considered by \tool are as follows.

\begin{itemize}
\item 
\emph{Multi-label classifiers} (meta): 
	BaggingML,
	BaggingMLdup,
	CM,
	DeepML,
	EM,
	EnsembleML,
	FilteredClassifier,
	MBR,
	MultiSearch,
		SubsetMapper,\\
	RandomSubspaceML

\item 
\emph{Multi-label classifiers} (base):
	Bayesian Classifier Chains (BCC),
	Back Propagation Neural Network (BPNN),
	Binary Relevance (BR),
	Binary Relevance quick (BRq),
	Classifier Chains (CC),
	Classifier Chains quick CCq,
	Conditional Dependency Networks (CDN),
	Conditional Dependency Trellis (CDT),
	Classifier Trellis (CT),
	Deep Back-Propagation Neural Network (DBPNN),
	Fourclass Pairwise (FW),
	Hierarchical Label Sets (HASEL),
	Label Combination (LC),
	Majority Label Set,
	Multi-lAbel classificatioN using AutoenCoders (Maniac),
	Monte-Carlo Classifier Chains (MCC),
	Probabilistic Classifier Chains (PCC),
	PMCC,
	Pruned Sets (PS),
	Pruned Sets with Threshold (PSt),
	RAndom k-labEl pruned sets (RAkEL),
	RAndom k-labEL Disjoint pruned sets (RAkELd),
	Ranking+Threshold (RT),
	Multi-Label Classification using Boolean Matrix Decomposition (MLCBMaD)
\item 
\emph{Single-label classifiers} (meta):
	AdaBoostM1, AdditiveRegression, AttributeSelectedClassifier, Bagging, ClassificationViaRegression, LogitBoost, MultiClassClassifier, RandomCommittee, RandomSubspace, Stacking,
	Vote
\item 
\emph{Single-label classifiers} (base):
	BayesNet, NaiveBayes, NaiveBayesMultinomial, Logistic, MultilayerPerceptron, SimpleLinearRegression, SimpleLogistic, SMO, VotedPerceptron, IBk, KStar, JRip, M5Rules, OneR, PART, ZeroR, DecisionStump, J48, LMT, M5P, RandomForest, RandomTree, REPTree
	
\end{itemize}

\tool adopts the node evaluation function of ML-Plan.
As explained above, this node evaluation function draws random completions and evaluates the corresponding completed pipelines on a validation set.
Every pipeline is validated several times using different splits of size 70\%/30\% of the data available for search to make the estimate more reliable.

While ML-Plan adopts 0/1-loss for computing the loss of a single pipeline, there are different loss measures for multi-label classification (cf. Section \ref{sec:problem}).
In principle, each of the losses could be used to guide the search; we took the F-measure (\ref{eq:fmeasure}), which we consider as one of the most meaningful MLC performance measures.
As different losses are known to be potentially competitive \cite{DBLP:journals/ml/DembczynskiWCH12},
it would also be possible to conduct a multi-objective search, which we consider as an interesting idea for future work.

\section{Experimental Evaluation}
\label{sec:evaluaton}
We evaluate \tool as introduced in the previous section on various datasets and for different timeouts.
Since there is no dedicated tool for automated multi-label classification publicly available to compare with, we evaluate \tool against the baselines as introduced in the following.
The implementation of \tool is publicly available\footnote{\url{https://github.com/fmohr/AILIbs}}.

\subsection{Baselines}
\label{ssec:baseline}
In our experimental study, we challenge \tool{} by two baselines in order to empirically show that it is \ifthenelse{\boolean{long}}{beyond simply trying out prominent well-performing MLC classifiers, }{}better than randomly traversing the search space\ifthenelse{\boolean{long}}{,}{} and beyond a simple reduction of current \automl{} tools for multi-class classification.
\ifthenelse{\boolean{long}}{The outlined properties result in a total of three baselines that are explained in the subsequent sections.}{To this end, we define the baselines as follows.}

\ifthenelse{\boolean{long}}{\subsubsection{Knowledgeable Expert (KE)}\label{baseline-expert}

While engineering machine learning applications, machine learning experts rely on experience gained in previous projects and often their intuition about the problem plays a major role. As already reported in \cite{feurer2015efficient}, there is a rather small subset of methods respectively algorithms which are usually performing substantially better than the remaining methods. Hence, an expert would rather start by trying one of these methods of which he knows that they usually perform well. The other way around, the expert will probably not try out models that perform inferior most of the time.

With this in mind, we define the first baseline to try out all possible combinations of all base multi-label classifiers with either RandomForest or SMO as a base learner as these base learners are known to perform particularly strong. In contrast to \tool and the other baselines, the knowledgeable expert baseline has no time constraints. This is due to reflect the advantage of the expert's intuition and experience gained from observed past performances.}{}

\subsubsection{Random Search (RS)}\label{baseline-random}

First, to demonstrate the effectiveness of \tool's strategy to search for good candidates, the second baseline is a random search which samples random candidates from the same search space.
We let \tool{} and the random search operate on the same search space, as for instance by optimizing the search space structure a bias towards better performing candidates could be introduced.
As a direct consequence, all the combinations possible for \tool{} are also available for the random search.
Thus, the only difference between \tool{} and the random search baseline is the strategy of traversing the search space.
The random search baseline is bound to the same timeout specification as \tool.

\subsubsection{Reduction to AutoML for Binary/Multinomial Classification (BR-AW)}\label{baseline-reduction}

Second, we define a baseline as an optimized version of Binary Relevance (BR), which is often used as a baseline in the multi-label classification literature. 
As BR reduces the MLC problem to a set of binary classification problems and the primary function of current \automl{} tools \cite{feurer2015efficient,autoweka,olson2016tpot,DBLP:conf/eurogp/SaPOP17} addresses binary respectively multinomial classification or regression problems, we leverage the well-established \automl{} tool \autoweka{} to individually optimize each of the base learners for the induced binary classification problems.
Note that \autoweka{} not only configures a classifier for the binary problems, but also an individualized pre-processing if appropriate.
To motivate this baseline, note that it would arguably be the most simple choice of an inexperienced user who is trying to incorporate \emph{some} automation into the process of solving an MLC problem. 

To guarantee a fair comparison in terms of permitted run time, we divided the available resources by the number of labels and, hence, the number of binary problems to be solved by \autoweka.
More precisely, given a timeout $t$ for \tool, the entire baseline algorithm should also have time $t$ available.
Consequently, we assigned $t / l$ as a timeout to each \autoweka{} instance within BR.
While it would  of course be possible to parallelize the learning process in BR and, hence, relax the timeout for each binary problem, no such parallelization is realized within MEKA.
\tool{} also adopts the MEKA implementation of BR and would not benefit from any parallelization. In our view, this setup therefore maximizes fairness for the comparison.

Since it might be the case that \autoweka{} fails to provide a result within the given time bound, two instances of BR using either SMO or Random Trees as base learners are evaluated in parallel for all labels. In these two cases, neither hyperparameter optimization nor preprocessing methods are configured.
The baseline then assumes the maximum over the three performances of \autoweka{} and the backup BR variants.

\subsection{Experimental Setup}
\label{sec:evaluaton:setup}
Results were obtained by carrying out 25 runs on 24 datasets with three different timeouts.
The datasets stem from the MULAN project website, but an independent copy is available\footnote{link hidden during review, MULAN sources are \url{http://mulan.sourceforge.net/datasets-mlc.html}}.
The significance of an improvement per dataset was determined using a t-test with a threshold for the p-value of 0.05.

We considered timeouts of one minute, one hour, and 12 hours.
Depending on the overall timeout, the timeout for the internal evaluation of a single solution by \tool{} was set to 10s for 1 minute runs and 5m for the other cases.
Runs that did not adhere to the time or resource limitations (plus a tolerance threshold) were canceled without considering their results.
That is, we canceled the algorithms if they did not terminate within 110\% of the predefined timeout.
Likewise, the algorithms were killed if they consumed more resources (memory or CPU) than allowed, which happens as both implementations fork new processes whose overall CPU and memory consumption is hard to control.

In each run, we used 70\% of a randomized split of the data for learning (search) and 30\% for testing.
We used the \emph{same} splits for all candidates, i.e., for each split and each timeout, we ran once the baseline and once \tool.
Note that there is no natural way of stratifying splits as in multi-class classification.
While there \emph{are} approaches to obtain such splits \cite{DBLP:conf/pkdd/SechidisTV11}, we used random splits for implementation-related reasons.
The relative performance of the algorithms should not be significantly affected by the split technique, since the same splits were used for all algorithms.

The computations were executed on (up to) 150 Linux machines in parallel, each of which with a resource limitation of 8 cores (Intel Xeon E5-2670, 2.6Ghz) and 32GB memory.
The total run-time was over 130k CPU hours (more than 14 CPU years).

\subsection{Results of Evaluation on Test Data}
\label{sec:evaluaton:results}
\newcommand{\impBRhour}{21}
\newcommand{\degBRhour}{2}
\newcommand{\impBRday}{20}
\newcommand{\degBRday}{1}
\newcommand{\impRSmin}{8}
\newcommand{\degRSmin}{5}
\newcommand{\impRShour}{12}
\newcommand{\degRShour}{2}
\newcommand{\impRSday}{14}
\newcommand{\degRSday}{2}

In Table~\ref{table:f-measure}, we report the results on the averaged (instance-wise) F-measure (\ref{eq:fmeasure}), for which \tool{} optimized. We compare \tool{} to the outlined baselines \ifthenelse{\boolean{long}}{Knowledgeable Expert (KE), }{}Random Search (RS)\ifthenelse{\boolean{long}}{, and}{ and} the reduction to \automl{} optimizing each single base learner of binary relevance with the help of \autoweka\ (BR-AW). Best results are highlighted in bold; significant improvements of \tool{} over a baseline according to the t-test is indicated by $\bullet$ and a significant degradation by $\circ$ (only comparing for the same timeout).
Results for using Auto-WEKA with binary relevance for a timeout of one minute do not exist---since the minimum run-time of Auto-WEKA is one minute, it cannot be used in that scenario.

The overall picture is that \tool{} dominates the baselines to a large extent.
Compared to BR-AW, \tool{} yields a significantly degraded result in \degBRhour{} of 24 cases for the 1h timeout and \degBRday/24 for the 24h timeout.
For most of the remaining datasets and both timeouts, the results returned by \tool{} significantly better than the ones returned by BR-AW.
In many data sets, even the results obtained within one minute are better than the ones returned by BR-AW for the 12h timeout.
Thus, it clearly outperforms a naive reduction of multi-label classification to binary relevance incorporating a single-label classification \automl{} tool.
The worse performance might be due to the handling of the timeout constraints for BR-AW as the overall run-time is divided by the number of labels.
However, BR-AW manages to perform significantly better than \tool{} within a timeout of 1h for the dataset with the second-most number of labels, which somehow contradicts this intuition.
Another reason is that although binary relevance is often already a very strong base for the comparison of single MLC strategies, it is clearly defeated by a portfolio of MLC strategies from which \tool{} is allowed to choose.
A consequence from this conjecture is that algorithm selection proves beneficial also for MLC.

Considering the random search baseline (RS), we observe that, for a one minute timeout, there is no clear winner or loser. While \tool{} achieves significant improvements over RS in \impRSmin{} cases, RS in turn yields superior results in \degRSmin{} cases. In a few cases RS does not manage to return any result within the given timeout as it is not bound to an internal timeout for evaluating candidates. This picture slowly changes when increasing the timeout to 1h, where \tool{} achieves \impRShour{} significant improvements over RS and only \degRShour{} significant degradations. This trend is continued when moving on to the timeout of 12h, where a significant degradation is obtained on only \degRSday{} datasets, while there are \impRSday{} significant improvements to be noted. Summing up the observations, we notice that for smaller timeouts \tool{} first behaves similar to a random search, which is somehow intuitive as some of the nodes have to be evaluated first before the search may notice high performance regions. Over time, \tool{} manages to find superior results as compared to the results found by the random search. From this, we conclude that the best-first search adopted by ML-Plan and also used in \tool{} to find well-performing multi-label classifiers proves beneficial for the \automl{} multi-label classification problem.

Based on these observations, a dedicated tool for multi-label classification can clearly be justified.
Investing into a guided search through the \tool{} search space brings a significant advantage over the proposed baselines.
We do not claim that combining reduction techniques with the use of standard \automl{} tools is not a meaningful approach---quite to the contrary.
However, instead of using a predefined reduction, the most appropriate one should be found in a systematic way, like in \tool.
The observations in the next section confirm that it is unlikely that there is a single-best multi-label classifier that could serve for this purpose.

In addition, the results after 12 hours are also equal or better than those reported on automated multi-label classification \cite{DBLP:conf/gecco/SaPF17}, despite the fact that the multi-label classifiers were trained on 90\% of a dataset and tested on 10\%.
In their workshop paper, de S\'a et.\ al evaluate their evolutionary approach on the three datasets {\sc Birds}, {\sc flags}, and {\sc Scene}.
We calculate that the considered timeout was roughly 52 CPU days, which would mean that, for a parallelization on a 16 core machine, this would be roughly 6 times as much time as the 12 hours granted for \tool\footnote{A notice for the reviewer: A direct experimental comparison was not possible, because de S\'a et al. did neither publish their code nor did they provide it (or binaries) on request.}.
As a consequence, there is currently no known tool that achieves better results on automated multi-label classification than \tool.

\begin{table}[t]
\resizebox{\textwidth}{!}{
\begin{tabular}{l|r|r|r||r|r|r|r|r|r|r|r}
\hline
\multirow{2}{*}{Dataset} &\multirow{2}{*}{\#Inst.} &\multirow{2}{*}{\#Att.} &\multirow{2}{*}{\#L} & \multicolumn{2}{c|}{1m}
& \multicolumn{3}{c|}{1h}
& \multicolumn{3}{c}{12h}\\
& & & 
& \multicolumn{1}{c|}{\tool} & \multicolumn{1}{c|}{RS} 
& \multicolumn{1}{c|}{\tool} & \multicolumn{1}{c|}{BR-AW} & \multicolumn{1}{c|}{RS}
& \multicolumn{1}{c|}{\tool} & \multicolumn{1}{c|}{BR-AW} & \multicolumn{1}{c}{RS}\\\hline

{\sc Arts} & 7484 & 23146 & 26 & \textbf{22.4$\pm$1.4} & 20.5$\pm$0.0 $\bullet$ & \textbf{50.1$\pm$8.3} & 28.0$\pm$0.8 $\bullet$ & 0.0$\pm$0.0 $\bullet$ & \textbf{53.0$\pm$1.2} & 28.0$\pm$0.8 $\bullet$ & 19.1$\pm$14 $\bullet$\\
{\sc Bibtex} & 7395 & 1836 & 159 & 10.3$\pm$2.9 & \textbf{17.1$\pm$6.1} $\circ$ & 33.0$\pm$0.5 & \textbf{39.6$\pm$0.7} $\circ$ & 22.6$\pm$2.2 $\bullet$ & 35.4$\pm$2.0 & \textbf{39.6$\pm$0.7} $\circ$ & 28.8$\pm$14 $\phantom{\circ}$\\
{\sc Birds} & 645 & 258 & 21 & \textbf{36.1$\pm$2.8} & 29.5$\pm$9.0 $\bullet$ & 37.4$\pm$3.2 & 32.8$\pm$2.5 $\bullet$ & \textbf{37.5$\pm$4.2} $\phantom{\circ}$ & \textbf{39.5$\pm$1.9} & 38.9$\pm$1.5 $\phantom{\circ}$ & 36.6$\pm$3.0 $\bullet$\\
{\sc Bookmarks} & 87856 & 2150 & 208 & \textbf{10.7$\pm$3.5} &  & \textbf{21.3$\pm$1.6} &  &  & \textbf{22.1$\pm$1.2} &  & \\
{\sc Business} & 11214 & 21924 & 30 & \textbf{73.0$\pm$0.7} &  & 73.9$\pm$2.6 & \textbf{75.5$\pm$0.9} $\circ$ & 73.1$\pm$0.4 $\phantom{\circ}$ & \textbf{79.7$\pm$0.5} & 75.5$\pm$0.9 $\bullet$ & 73.5$\pm$0.3 $\bullet$\\
{\sc Computers} & 12444 & 34096 & 33 & 44.0$\pm$0.7 & \textbf{51.7$\pm$7.4} $\circ$ & \textbf{46.8$\pm$5.9} & 37.3$\pm$0.6 $\bullet$ & 44.3$\pm$0.0 $\phantom{\circ}$ & \textbf{64.0$\pm$0.7} & 37.3$\pm$0.6 $\bullet$ & 42.2$\pm$9.6 $\bullet$\\
{\sc Education} & 12030 & 27534 & 33 & 27.7$\pm$0.6 & \textbf{51.5$\pm$0.0} $\circ$ & \textbf{48.1$\pm$10} & 24.8$\pm$0.7 $\bullet$ & 38.4$\pm$14 $\bullet$ & \textbf{53.1$\pm$0.6} & 24.8$\pm$0.7 $\bullet$ & 20.2$\pm$14 $\bullet$\\
{\sc Enron} & 1702 & 1001 & 53 & \textbf{46.6$\pm$3.5} & 39.6$\pm$9.3 $\bullet$ & \textbf{54.1$\pm$1.0} & 50.9$\pm$1.1 $\bullet$ & 48.5$\pm$6.4 $\bullet$ & 56.3$\pm$0.7 & 46.6$\pm$9.7 $\bullet$ & \textbf{57.5$\pm$2.8} $\phantom{\circ}$\\
{\sc Entertainment} & 12730 & 32001 & 21 & 26.7$\pm$1.9 & \textbf{32.5$\pm$0.0} $\circ$ & \textbf{40.6$\pm$15} & 32.9$\pm$0.6 $\bullet$ & 31.8$\pm$16 $\phantom{\circ}$ & \textbf{67.4$\pm$0.5} & 32.9$\pm$0.6 $\bullet$ & 37.9$\pm$13 $\bullet$\\
{\sc Flags} & 194 & 14 & 12 & 64.3$\pm$3.2 & \textbf{66.4$\pm$3.2} $\circ$ & 67.0$\pm$1.8 & 65.2$\pm$2.8 $\bullet$ & \textbf{68.5$\pm$2.9} $\circ$ & 66.3$\pm$1.5 & 65.6$\pm$1.3 $\phantom{\circ}$ & \textbf{68.6$\pm$2.1} $\circ$\\
{\sc Health} & 9205 & 30605 & 32 & \textbf{42.8$\pm$1.0} & 41.7$\pm$0.9 $\bullet$ & \textbf{62.5$\pm$14} & 55.3$\pm$0.8 $\bullet$ & 33.4$\pm$8.8 $\bullet$ & \textbf{72.9$\pm$1.1} & 55.3$\pm$0.8 $\bullet$ & 48.1$\pm$4.0 $\bullet$\\
{\sc LangLog} & 1460 & 1004 & 75 & \textbf{8.8$\pm$3.8} & 6.5$\pm$4.2 $\phantom{\circ}$ & \textbf{19.9$\pm$1.8} & 13.8$\pm$1.1 $\bullet$ & 11.5$\pm$5.1 $\bullet$ & \textbf{19.7$\pm$0.6} & 13.8$\pm$1.1 $\bullet$ & 12.7$\pm$4.2 $\bullet$\\
{\sc MEDC} & 978 & 1449 & 45 & \textbf{78.0$\pm$1.6} & 64.9$\pm$17 $\bullet$ & \textbf{78.8$\pm$1.4} & 77.3$\pm$1.3 $\bullet$ & 77.7$\pm$3.7 $\phantom{\circ}$ & 77.5$\pm$2.6 & 72.2$\pm$4.1 $\bullet$ & \textbf{79.2$\pm$2.3} $\circ$\\
{\sc Mediamill} & 43907 & 120 & 101 & \textbf{38.9$\pm$12} &  & \textbf{56.1$\pm$4.1} & 49.4$\pm$0.2 $\bullet$ & 44.6$\pm$0.0 $\bullet$ & \textbf{60.0$\pm$1.5} & 49.4$\pm$0.2 $\bullet$ & 48.6$\pm$5.1 $\bullet$\\
{\sc Musicout} & 593 & 72 & 6 & \textbf{66.2$\pm$3.0} & 62.9$\pm$5.1 $\bullet$ & \textbf{67.0$\pm$3.1} & 57.8$\pm$3.6 $\bullet$ & 66.2$\pm$2.7 $\phantom{\circ}$ & \textbf{68.1$\pm$1.1} & 60.2$\pm$1.5 $\bullet$ & 67.6$\pm$2.1 $\phantom{\circ}$\\
{\sc Protein} & 662 & 1186 & 27 & \textbf{99.0$\pm$0.4} & 73.3$\pm$36 $\bullet$ & \textbf{99.0$\pm$0.5} & 58.2$\pm$0.0 $\bullet$ & 97.9$\pm$2.8 $\phantom{\circ}$ & \textbf{98.7$\pm$0.4} & 79.1$\pm$18 $\bullet$ & \textbf{98.7$\pm$0.7} $\phantom{\circ}$\\
{\sc Recreation} & 12828 & 30324 & 22 & \textbf{20.8$\pm$3.1} &  & \textbf{35.5$\pm$18} & 26.9$\pm$1.0 $\bullet$ & 30.9$\pm$26 $\phantom{\circ}$ & \textbf{63.1$\pm$0.6} & 26.9$\pm$1.0 $\bullet$ & 33.9$\pm$5.5 $\bullet$\\
{\sc Reference} & 8027 & 39679 & 33 & 44.1$\pm$0.7 & \textbf{44.3$\pm$0.0} $\phantom{\circ}$ & \textbf{54.2$\pm$9.7} & 49.0$\pm$0.4 $\bullet$ & 48.4$\pm$5.3 $\bullet$ & \textbf{63.5$\pm$0.7} & 49.0$\pm$0.4 $\bullet$ & 52.0$\pm$0.0 $\bullet$\\
{\sc Scene} & 2407 & 294 & 6 & \textbf{62.7$\pm$2.4} & 57.1$\pm$19 $\phantom{\circ}$ & \textbf{76.1$\pm$1.1} & 63.0$\pm$2.0 $\bullet$ & 72.3$\pm$4.4 $\bullet$ & \textbf{77.6$\pm$0.8} & 63.3$\pm$1.4 $\bullet$ & 77.1$\pm$2.1 $\phantom{\circ}$\\
{\sc Science} & 6428 & 37187 & 40 & \textbf{22.0$\pm$1.1} & 21.8$\pm$0.0 $\phantom{\circ}$ & 45.9$\pm$16 & 29.4$\pm$0.9 $\bullet$ & \textbf{53.6$\pm$0.0} $\circ$ & \textbf{56.3$\pm$0.5} & 29.4$\pm$0.9 $\bullet$ & 16.3$\pm$21 $\bullet$\\
{\sc Social} & 12111 & 52350 & 39 & \textbf{41.8$\pm$3.5} & 2.6$\pm$0.0 $\bullet$ & \textbf{47.4$\pm$10} & 38.9$\pm$2.1 $\bullet$ & 0.0$\pm$0.0 $\bullet$ & \textbf{68.1$\pm$0.6} & 38.9$\pm$2.1 $\bullet$ & 55.3$\pm$3.4 $\bullet$\\
{\sc Society} & 14512 & 31802 & 27 & \textbf{39.7$\pm$1.5} &  & \textbf{43.3$\pm$6.5} & 30.2$\pm$0.4 $\bullet$ & 19.9$\pm$16 $\bullet$ & \textbf{55.6$\pm$0.9} & 30.2$\pm$0.4 $\bullet$ & \\
{\sc Tmc} & 28596 & 49060 & 22 & \textbf{20.5$\pm$0.4} &  & \textbf{35.4$\pm$1.5} &  & 33.7$\pm$0.5 $\bullet$ &  &  & \textbf{34.7$\pm$0.0}\\
{\sc Yeast} & 2417 & 103 & 14 & \textbf{60.6$\pm$1.4} & 56.8$\pm$11 $\phantom{\circ}$ & \textbf{64.4$\pm$0.8} & 60.1$\pm$2.1 $\bullet$ & 64.0$\pm$2.1 $\phantom{\circ}$ & \textbf{65.2$\pm$0.9} & 61.3$\pm$1.2 $\bullet$ & 64.1$\pm$1.6 $\bullet$\\


\hline
\end{tabular}}
\caption{Means and standard deviation of instance-wise F-Measure. Each entry represents the mean and standard-deviation over 25 runs with different random seeds. Missing values are due to returning no result in the given timeout or due to memory overflows which occurred in long runs on memory-intensive datasaets.}

\label{table:f-measure}
\end{table}

\subsection{Overview of Chosen Classifiers}
\label{sec:evaluaton:choices}
\begin{figure*}[ht]
\includegraphics[width=\textwidth]{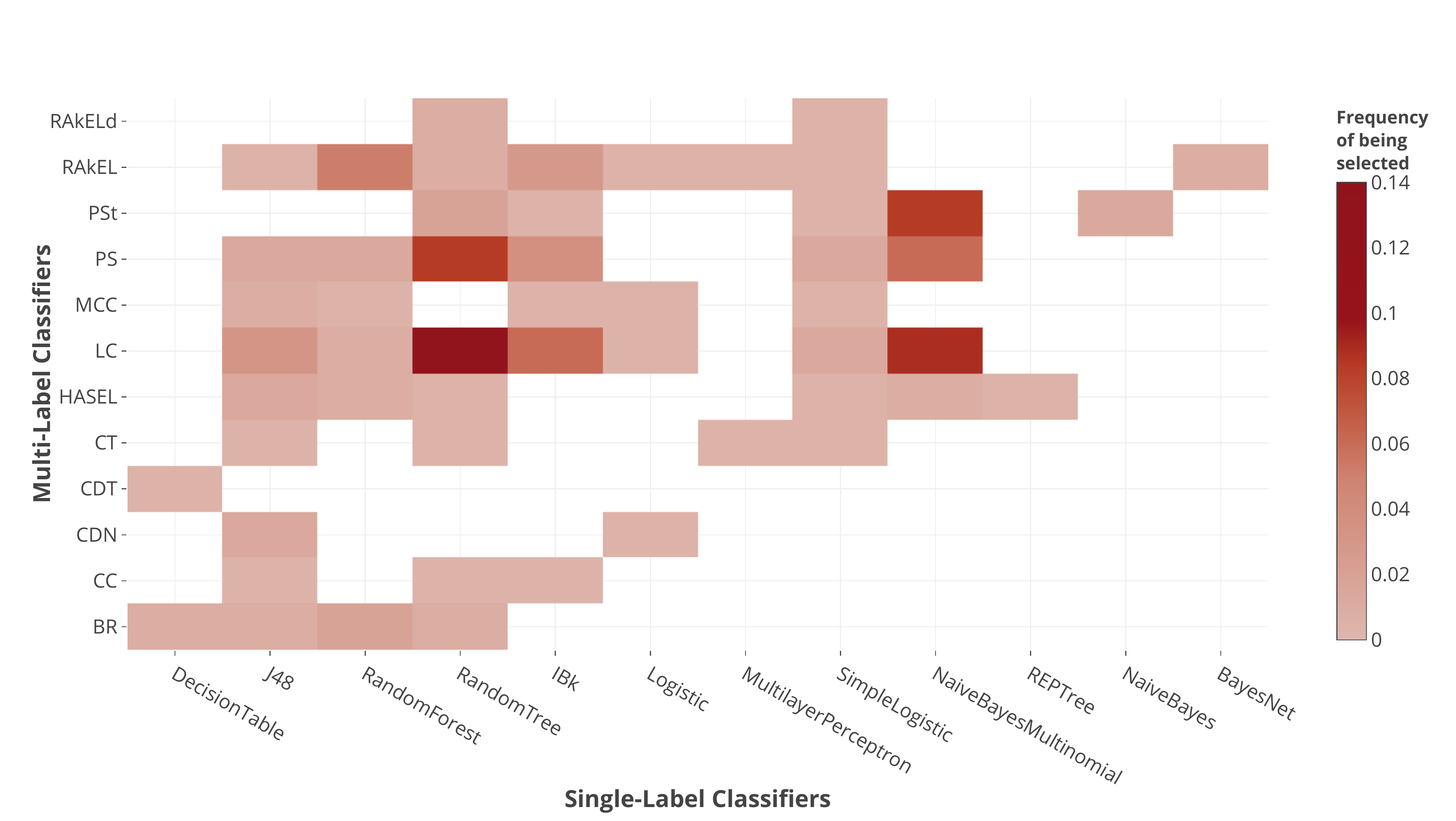}
\caption{Heatmap for selected combinations of multi-label classifier and base classifier}
\label{fig:classifiers}
\end{figure*}
The map in Fig. \ref{fig:classifiers} gives an overview of the algorithm combinations chosen by \tool{} for the scenario of a 1h timeout.
The vertical axis iterates over the multi-label classifiers, and the horizontal axis over the WEKA base classifiers plugged into the multi-label classifiers by \tool.
Note that the Majority Label Set classifier is not considered in this map, because it is not parametrized with a base learner; still, it was also selected a couple of times.
It is not surprising that the map looks rather sparse, because it is unlikely that every \emph{combination} of multi-label classifier and base classifier occurs.

The diversity of choices supports the idea of optimizing multi-label classification pipelines in an instance-specific manner instead of using a single multi-label classifier that is best on average.
We observe that each multi-label classifier was selected at least once and that, in general, there is no single classifier dominating the others.
Likewise, quite some variety of base learners was selected by \tool, and there is not really a dominant base learner\footnote{Random trees and Naive Bayes Multinomial are selected over average, but this is mainly due to their priority assigned by \tool{} during search as we explain in the following paragraph.}.
In fact, even if we focus on a single multi-label classifier, there are no specifically dominant base classifiers.

The observation that not all base learners were chosen at least once may be explained by the fact that not even all of them were \emph{tried} at least once.
Recalling the procedure of \tool{} sketched in Fig. \ref{fig:decomposition-hops}, we see that the base learner is the third algorithm choice, and \tool{} inherits an ordering among these learners from ML-Plan, in which they are analyzed.
It is clear that a base classifier with low rank will not be chosen by \tool if it has not even been tried before the timeout triggers.

Of course, this last point raises the question of scalability, which affects not only \tool but Auto-ML tools that evaluate candidates in general.
On one hand, it can be certainly said that these executing approaches, i.e., the approaches that execute a significant number of candidates during search (including \tool), do not scale with the data size.
For the case of automated multi-label classification, this is even more severe, since the search space is even more complex than for the ``simple'' case of multi-class classification (which itself is already infeasible).
On the other hand, there is currently no alternative in sight that would offer a more efficient solution.
Indeed, the algorithm selection community is aware of this and has adopted ideas to exploit knowledge gained in the past or during the search in order to avoid less promising evaluations \cite{DBLP:conf/ecai/KadiogluMST10,hutter2011sequential}.
However, at the present time, there is no smart solution in sight that relieves us from execution.

\section{Conclusion}
In this paper, we have presented \tool, an \automl{} tool to automatically select and configure a multi-label classifier for a given dataset.
On the model level, \tool{} adopts hierarchical task network (HTN) planning, a technique from AI planning, to recursively build multi-label classifiers.
On the code level, \tool{} binds these multi-label classifiers to implementations in the multi-label framework MEKA in order to execute them and evaluate their performance.
We demonstrate the usefulness of a tool dedicated to multi-label classification by comparing it to a random search baseline and a baseline which reduces a multi-label classification problem to binary relevance incorporating \autoweka{} to tailor each individual base learner. We show that, given a suitable timeout, \tool{} significantly outperforms both baselines.
To the best of our knowledge, apart from \cite{DBLP:conf/gecco/SaPF17,DBLP:conf/ppsn/SaFP18}, this is the first substantially evaluated approach to automated multi-label classification.

Having confirmed the suitability of a dedicated tool for automated multi-label classification, some natural follow-up research questions arise.
The most important question is associated with the issue of scalability, which affects almost all \automl{} frameworks, but which is particularly severe for multi-label classification due to an even more complex search space.
One possible way for scaling \tool{} would be to use it in a service-oriented architecture as proposed in \cite{mlsplan,mlsplanFull}.
Another question is what benefit can be expected of MLC pipelines incorporating preprocessing and being more flexible in both the configuration of parameters and the configuration of base classifiers decomposing the original problem into sub-problems, as proposed in \cite{reductionStumps} for multi-class classification, such that finally a heuristic strategy for HOMER \cite{homer} is obtained. Furthermore, building ensembles of optimized hierarchical decompositions could lead to further improvements of the overall performance, as shown for multi-class classification in \cite{evolvedNDs}.

This question is contrarious to the first one in the sense that any step in that direction will enlarge the search space even further.
Finally, the existence of different MLC loss measures motivates a multi-objective optimization process that not only considers a single measure but several such losses simultaneously.

\section*{Acknowledgements}
This work was partially supported by the German Research Foundation (DFG) within the Collaborative Research Center "On-The-Fly Computing" (SFB 901).
%
%
%
\bibliographystyle{splncs04}
\bibliography{literature}
%
%
%
%
%
\end{document}